# Machine Learning that Matters

**Kiri L. Wagstaff**  KIRI.L.WAGSTAFF@JPL.NASA.GOV

Jet Propulsion Laboratory, California Institute of Technology, 4800 Oak Grove Drive, Pasadena, CA 91109 USA

## Abstract

Much of current machine learning (ML) research has lost its connection to problems of import to the larger world of science and society. From this perspective, there exist glaring limitations in the data sets we investigate, the metrics we employ for evaluation, and the degree to which results are communicated back to their originating domains. What changes are needed to how we conduct research to increase the impact that ML has? We present six Impact Challenges to explicitly focus the field's energy and attention, and we discuss existing obstacles that must be addressed. We aim to inspire ongoing discussion and focus on ML that matters.

## 1. Introduction

At one time or another, we all encounter a friend, spouse, parent, child, or concerned citizen who, upon learning that we work in machine learning, wonders "What's it good for?" The question may be phrased more subtly or elegantly, but no matter its form, it gets at the motivational underpinnings of the work that we do. Why do we invest years of our professional lives in machine learning research? What difference does it make, to ourselves and to the world at large?

Much of machine learning (ML) research is inspired by weighty problems from biology, medicine, finance, astronomy, etc. The growing area of computational sustainability (Gomes, 2009) seeks to connect ML advances to real-world challenges in the environment, economy, and society. The CALO (Cognitive Assistant that Learns and Organizes) project aimed to integrate learning and reasoning into a desktop assistant, potentially impacting everyone who uses a computer (SRI International, 2003–2009). Machine learning has effectively solved spam email detection (Zdziarski, 2005) and machine translation (Koehn et al., 2003), two problems of global import. And so on.

And yet we still observe a proliferation of published ML papers that evaluate new algorithms on a handful of isolated benchmark data sets. Their "real world" experiments may operate on data that originated in the real world, but the results are rarely communicated back to the origin. Quantitative improvements in performance are rarely accompanied by an assessment of whether those gains matter to the world outside of machine learning research.

This phenomenon occurs because there is no widespread emphasis, in the training of graduate student researchers or in the review process for submitted papers, on connecting ML advances back to the larger world. Even the rich assortment of applications-driven ML research often fails to take the final step to translate results into impact.

Many machine learning problems are phrased in terms of an objective function to be optimized. It is time for us to ask a question of larger scope: what is the field's objective function? Do we seek to maximize performance on isolated data sets? Or can we characterize progress in a more meaningful way that measures the concrete impact of machine learning innovations?

This short position paper argues for a change in how we view the relationship between machine learning and science (and the rest of society). This paper does not contain any algorithms, theorems, experiments, or results. Instead it seeks to stimulate creative thought and research into a large but relatively unaddressed issue that underlies much of the machine learning field. The contributions of this work are 1) the clear identification and description of a fundamental problem: the frequent lack of connection between machine learning research and the larger world of scientific inquiry and humanity, 2) suggested first steps towards addressing this gap, 3) the issuance of relevant Impact Challenges to the machine learning community, and 4) the identification of several key obstacles to machine learning

---

impact, as an aid for focusing future research efforts. Whether or not the reader agrees with all statements in this paper, if it inspires thought and discussion, then its purpose has been achieved.

## 2. Machine Learning for Machine Learning's Sake

This section highlights aspects of the way ML research is conducted today that limit its impact on the larger world. Our goal is not to point fingers or critique individuals, but instead to initiate a critical self-inspection and constructive, creative changes. These problems do not trouble *all* ML work, but they are common enough to merit our effort in eliminating them.

The argument here is also not about "theory versus applications." Theoretical work can be as inspired by real problems as applied work can. The criticisms here focus instead on the limitations of work that lies between theory and meaningful applications: algorithmic advances accompanied by empirical studies that are divorced from true impact.

### 2.1. Hyper-Focus on Benchmark Data Sets

Increasingly, ML papers that describe a new algorithm follow a standard evaluation template. After presenting results on synthetic data sets to illustrate certain aspects of the algorithm's behavior, the paper reports results on a collection of standard data sets, such as those available in the UCI archive (Frank & Asuncion, 2010). A survey of the 152 non-cross-conference papers published at ICML 2011 reveals:

| | |
|---|---|
| 148/152 (93%) | include experiments of some sort |
| 57/148 (39%) | use synthetic data |
| 55/148 (37%) | use UCI data |
| 34/148 (23%) | use *ONLY* UCI and/or synthetic data |
| **1/148 (1%)** | **interpret results in domain context** |

The possible advantages of using familiar data sets include 1) enabling direct empirical comparisons with other methods and 2) greater ease of interpreting the results since (presumably) the data set properties have been widely studied and understood. However, in practice direct comparisons fail because we have no standard for reproducibility. Experiments vary in methodology (train/test splits, evaluation metrics, parameter settings), implementations, or reporting. Interpretations are almost never made. Why is this?

First, meaningful interpretations are hard. Virtually none of the ML researchers who work with these data sets happen to also be experts in the relevant scientific disciplines. Second, and more insidiously, the ML field neither motivates nor requires such interpretation. Reviewers do not inquire as to which classes were well classified and which were not, what the common error types were, or even why the particular data sets were chosen. There is no expectation that the authors report whether an observed $x\%$ improvement in performance promises any real impact for the original domain. Even when the authors have forged a collaboration with qualified experts, little paper space is devoted to interpretation, because we (as a field) do not require it.

The UCI archive has had a tremendous impact on the field of machine learning. Legions of researchers have chased after the best iris or mushroom classifier. Yet this flurry of effort does not seem to have had any impact on the fields of botany or mycology. Do scientists in these disciplines even need such a classifier? Do they publish about this subject in their journals?

There is not even agreement in the community about what role the UCI data sets serve (benchmark? "real-world"?). They are of less utility than synthetic data, since we did not control the process that generated them, and yet they fail to serve as real world data due to their disassociation from any real world context (experts, users, operational systems, etc.). It is as if we have forgotten, or chosen to ignore, that each data set is more than just a matrix of numbers. Further, the existence of the UCI archive has tended to over-emphasize research effort on classification and regression problems, at the expense of other ML problems (Langley, 2011). Informal discussions with other researchers suggest that it has also de-emphasized the need to learn how to formulate problems and define features, leaving young researchers unprepared to tackle new problems.

This trend has been going on for at least 20 years. Jaime Carbonell, then editor of *Machine Learning*, wrote in 1992 that "the standard Irvine data sets are used to determine percent accuracy of concept classification, without regard to performance on a larger external task" (Carbonell, 1992). Can we change that trend for the next 20 years? Do we want to?

### 2.2. Hyper-Focus on Abstract Metrics

There are also problems with how we measure performance. Most often, an abstract evaluation metric (classification accuracy, root of the mean squared error or RMSE, F-measure (van Rijsbergen, 1979), etc.) is used. These metrics are abstract in that they explicitly ignore or remove problem-specific details, usually so that numbers can be compared across domains. Does this seemingly obvious strategy provide us with useful information?

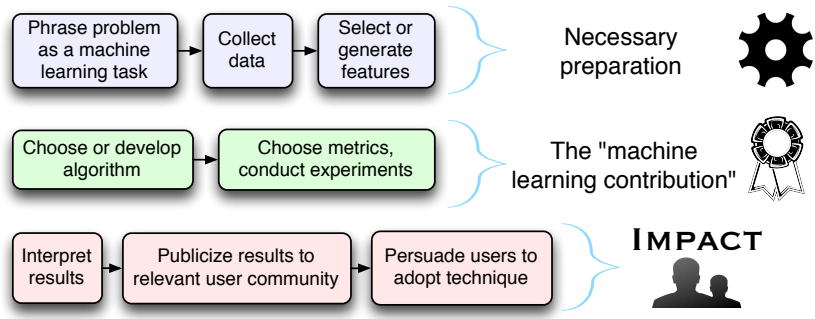

Figure 1. Three stages of a machine learning research program. Current publishing incentives are highly biased towards the middle row only.

It is recognized that the performance obtained by training a model $\mathcal{M}$ on data set $X$ may not reflect $\mathcal{M}$'s performance on other data sets drawn from the same problem, i.e., training loss is an underestimate of test loss (Hastie et al., 2001). Strategies such as splitting $X$ into training and test sets or cross-validation aim to estimate the expected performance of $\mathcal{M}'$, trained on *all* of $X$, when applied to future data $X'$.

However, these metrics tell us nothing about the *impact* of different performance. For example, 80% accuracy on iris classification might be sufficient for the botany world, but to classify as poisonous or edible a mushroom you intend to ingest, perhaps 99% (or higher) accuracy is required. The assumption of cross-domain comparability is a mirage created by the application of metrics that have the same range, but not the same meaning. Suites of experiments are often summarized by the average accuracy across all data sets. This tells us *nothing at all useful* about generalization or impact, since the meaning of an $x\%$ improvement may be very different for different data sets. A related problem is the persistence of "bake-offs" or "mindless comparisons among the performance of algorithms that reveal little about the sources of power or the effects of domain characteristics" (Langley, 2011).

Receiver Operating Characteristic (ROC) curves are used to describe a system's behavior for a range of threshold settings, but they are rarely accompanied by a discussion of which performance regimes are relevant to the domain. The common practice of reporting the area under the curve (AUC) (Hanley & McNeil, 1982) has several drawbacks, including summarizing performance over all possible regimes even if they are unlikely ever to be used (e.g., extremely high false positive rates), and weighting false positives and false negatives equally, which may be inappropriate for a given problem domain (Lobo et al., 2008). As such, it is insufficiently grounded to meaningfully measure impact.

Methods from statistics such as the t-test (Student, 1908) are commonly used to support a conclusion about whether a given performance improvement is "significant" or not. Statistical significance is a function of a set of numbers; it does not compute real-world significance. Of course we all *know* this, but it rarely inspires the addition of a separate measure of (true) significance. How often, instead, a t-test result serves as the final punctuation to an experimental utterance!

### 2.3. Lack of Follow-Through

It is easy to sit in your office and run a Weka (Hall et al., 2009) algorithm on a data set you downloaded from the web. It is very hard to identify a problem for which machine learning may offer a solution, determine what data should be collected, select or extract relevant features, choose an appropriate learning method, select an evaluation method, interpret the results, involve domain experts, publicize the results to the relevant scientific community, persuade users to adopt the technique, and (only then) to truly have made a difference (see Figure 1). An ML researcher might well feel fatigued or daunted just contemplating this list of activities. However, each one is a necessary component of any research program that seeks to have a real impact on the world outside of machine learning.

Our field imposes an additional obstacle to impact. Generally speaking, only the activities in the middle row of Figure 1 are considered "publishable" in the ML community. The Innovative Applications of Artificial Intelligence conference and International Conference on Machine Learning and Applications are exceptions. The International Conference on Machine Learning (ICML) experimented with an (unreviewed) "invited applications" track in 2010. Yet to be accepted as a mainstream paper at ICML or *Machine Learning* or the *Journal of Machine Learning Research*, authors must demonstrate a "machine learning contribution" that is often narrowly interpreted by reviewers as "the development of a new algorithm or the explication of a novel theoretical analysis." While these are excellent, laudable advances, unless there is an equal expectation for the bottom row of Figure 1, there is little incentive to connect these advances with the outer world.

Reconnecting active research to relevant real-world problems is part of the process of maturing as a research field. To the rest of the world, these are the only visible advances of ML, its only contributions to the larger realm of science and human endeavors.

## 3. Making Machine Learning Matter

Rather than following the letter of machine learning, can we reignite its spirit? This is not simply a matter of reporting on isolated applications. What is needed is a fundamental change in how we formulate, attack, and evaluate machine learning research projects.

### 3.1. Meaningful Evaluation Methods

The first step is to define or select evaluation methods that enable direct measurement, wherever possible, of the impact of ML innovations. In addition to traditional measures of performance, we can measure dollars saved, lives preserved, time conserved, effort reduced, quality of living increased, and so on. Focusing our metrics on impact will help motivate upstream restructuring of research efforts. They will guide how we select data sets, structure experiments, and define objective functions. At a minimum, publications can report how a given improvement in accuracy translates to impact for the originating problem domain.

The reader may wonder how this can be accomplished, if our goal is to develop general methods that apply across domains. Yet (as noted earlier) the common approach of using the same metric for all domains relies on an unstated, and usually unfounded, assumption that it is possible to equate an $x\%$ improvement in one domain with that in another. Instead, if the same method can yield profit improvements of $10,000 per year for an auto-tire business as well as the avoidance of 300 unnecessary surgical interventions per year, then it will have demonstrated a powerful, wide-ranging utility.

### 3.2. Involvement of the World Outside ML

Many ML investigations involve domain experts as collaborators who help define the ML problem and label data for classification or regression tasks. They can also provide the missing link between an ML performance plot and its significance to the problem domain. This can help reduce the number of cases where an ML system perfectly solves a sub-problem of little interest to the relevant scientific community, or where the ML system's performance appears good numerically but is insufficiently reliable to ever be adopted.

We could also solicit short "Comment" papers, to accompany the publication of a new ML advance, that are authored by researchers with relevant domain expertise but who were uninvolved with the ML research. They could provide an independent assessment of the performance, utility, and impact of the work. As an additional benefit, this informs new communities about how, and how well, ML methods work. Raising awareness, interest, and buy-in from ecologists, astronomers, legal experts, doctors, etc., can lead to greater opportunities for machine learning impact.

### 3.3. Eyes on the Prize

Finally, we should consider potential impact when selecting which research problems to tackle, not merely how interesting or challenging they are from the ML perspective. How many people, species, countries, or square meters would be impacted by a solution to the problem? What level of performance would constitute a meaningful improvement over the status quo?

Warrick et al. (2010) provides an example of ML work that tackles all three aspects. Working with doctors and clinicians, they developed a system to detect fetal hypoxia (oxygen deprivation) and enable emergency intervention that literally saves babies from brain injuries or death. After publishing their results, which demonstrated the ability to have detected 50% of fetal hypoxia cases early enough for intervention, with an acceptable false positive rate of 7.5%, they are currently working on clinical trials as the next step towards wide deployment. Many such examples exist. This paper seeks to inspire more.

## 4. Machine Learning Impact Challenges

One way to direct research efforts is to articulate ambitious and meaningful challenges. In 1992, Carbonell articulated a list of challenges for the field, not to increase its impact but instead to "put the fun back into machine learning" (Carbonell, 1992). They included:

1. Discovery of a new physical law leading to a published, referred scientific article.

2. Improvement of 500 USCF/FIDE chess rating points over a class B level start.

3. Improvement in planning performance of 100 fold in two different domains.

4. Investment earnings of $1M in one year.

5. Outperforming a hand-built NLP system on a task such as translation.
6. Outperforming all hand-built medical diagnosis systems with an ML solution that is deployed and regularly used at at least two institutions.

Because impact was not the guiding principle, these challenges range widely along that axis. An improved chess player might arguably have the lowest real-world impact, while a medical diagnosis system in active use could impact many human lives.

We therefore propose the following six Impact Challenges as examples of machine learning that matters:

1. A law passed or legal decision made that relies on the result of an ML analysis.
2. $100M saved through improved decision making provided by an ML system.
3. A conflict between nations averted through high-quality translation provided by an ML system.
4. A 50% reduction in cybersecurity break-ins through ML defenses.
5. A human life saved through a diagnosis or intervention recommended by an ML system.
6. Improvement of 10% in one country's Human Development Index (HDI) (Anand & Sen, 1994) attributable to an ML system.

These challenges seek to capture the entire process of a successful machine learning endeavor, including performance, infusion, and impact. They differ from existing challenges such as the DARPA Grand Challenge (Buehler et al., 2007), the Netflix Prize (Bennett & Lanning, 2007), and the Yahoo! Learning to Rank Challenge (Chapelle & Chang, 2011) in that they do not focus on any single problem domain, nor a particular technical capability. The goal is to inspire the field of machine learning to take the steps needed to mature into a valuable contributor to the larger world.

No such list can claim to be comprehensive, including this one. It is hoped that readers of this paper will be inspired to formulate additional Impact Challenges that will benefit the entire field.

Much effort is often put into chasing after goals in which an ML system outperforms a human at the same task. The Impact Challenges in this paper also differ from that sort of goal in that human-level performance is not the gold standard. What matters is achieving performance sufficient to make an impact on the world. As an analogy, consider a sick child in a rural setting. A neighbor who runs two miles to fetch the doctor need not achieve Olympic-level running speed (performance), so long as the doctor arrives in time to address the sick child's needs (impact).

## 5. Obstacles to ML Impact

Let us imagine a machine learning researcher who is motivated to tackle problems of widespread interest and impact. What obstacles to success can we foresee? Can we set about eliminating them in advance?

**Jargon.** This issue is endemic to all specialized research fields. Our ML vocabulary is so familiar that it is difficult even to detect when we're using a specialized term. Consider a handful of examples: "feature extraction," "bias-variance tradeoff," "ensemble methods," "cross-validation," "low-dimensional manifold," "regularization," "mutual information," and "kernel methods." These are all basic concepts within ML that create conceptual barriers when used glibly to communicate with others. Terminology can serve as a barrier not just for domain experts and the general public but even between closely related fields such as ML and statistics (van Iterson et al., 2012). We should explore and develop ways to express the same ideas in more general terms, or even better, in terms already familiar to the audience. For example, "feature extraction" can be termed "representation;" the notion of "variance" can be "instability;" "cross-validation" is also known as "rotation estimation" outside of ML; "regularization" can be explained as "choosing simpler models;" and so on. These terms are not as precise, but more likely to be understood, from which a conversation about further subtleties can ensue.

**Risk.** Even when an ML system is no more, or less, prone to error than a human performing the same task, relying on the machine can feel riskier because it raises new concerns. When errors are made, where do we assign culpability? What level of ongoing commitment do the ML system designers have for adjustments, upgrades, and maintenance? These concerns are especially acute for fields such as medicine, spacecraft, finance, and real-time systems, or exactly those settings in which a large impact is possible. An increased sphere of impact naturally also increases the associated risk, and we must address those concerns (through technology, education, and support) if we hope to infuse ML into real systems.

**Complexity.** Despite the proliferation of ML toolboxes and libraries, the field has not yet matured to a point where researchers from other areas can simply apply ML to the problem of their choice (as they might do with methods from physics, math, mechanical engineering, etc.). Attempts to do so often fail due to lack of knowledge about how to phrase the problem, what features to use, how to search over parameters, etc. (i.e., the top row of Figure 1). For this reason, it has been said that ML solutions come "packaged in a Ph.D.;" that is, it requires the sophistication of a grad-uate student or beyond to successfully deploy ML to solve real problems—and that same Ph.D. is needed to maintain and update the system after its deployment. It is evident that this strategy does not scale to the goal of widespread ML impact. Simplifying, maturing, and robustifying ML algorithms and tools, while itself an abstract activity, can help erode this obstacle and permit wider, independent uses of ML.

## 6. Conclusions

Machine learning offers a cornucopia of useful ways to approach problems that otherwise defy manual solution. However, much current ML research suffers from a growing detachment from those real problems. Many investigators withdraw into their private studies with a copy of the data set and work in isolation to perfect algorithmic performance. Publishing results to the ML community is the end of the process. Successes usually are not communicated back to the original problem setting, or not in a form that can be used.

Yet these opportunities for real impact are widespread. The worlds of law, finance, politics, medicine, education, and more stand to benefit from systems that can analyze, adapt, and take (or at least recommend) action. This paper identifies six examples of Impact Challenges and several real obstacles in the hope of inspiring a lively discussion of how ML can best make a difference. Aiming for real impact does not just increase our job satisfaction (though it may well do that); it is the only way to get the rest of the world to notice, recognize, value, and adopt ML solutions.

## Acknowledgments

We thank Tom Dietterich, Terran Lane, Baback Moghaddam, David Thompson, and three insightful anonymous reviewers for suggestions on this paper. This work was performed while on sabbatical from the Jet Propulsion Laboratory.

## References

Anand, Sudhir and Sen, Amartya K. Human development index: Methodology and measurement. Human Development Report Office, 1994.

Bennett, James and Lanning, Stan. The Netflix Prize. In *Proc. of KDD Cup and Workshop*, pp. 3–6, 2007.

Buehler, Martin, Iagnemma, Karl, and Singh, Sanjiv (eds.). *The 2005 DARPA Grand Challenge: The Great Robot Race*. Springer, 2007.

Carbonell, Jaime. Machine learning: A maturing field. *Machine Learning*, 9:5–7, 1992.

Chapelle, Olivier and Chang, Yi. Yahoo! Learning to Rank Challenge overview. *JMLR: Workshop and Conference Proceedings*, 14, 2011.

Frank, A. and Asuncion, A. UCI machine learning repository, 2010. URL http://archive.ics.uci.edu/ml.

Gomes, Carla P. Computational sustainability: Computational methods for a sustainable environment, economy, and society. *The Bridge*, 39(4):5–13, Winter 2009. National Academy of Engineering.

Hall, M., Frank, E., Holmes, G., Pfahringer, B., Reutemann, P., and Witten, I. H. The WEKA data mining software: An update. *SIGKDD Explorations*, 11(1):10–18, 2009.

Hanley, J. A. and McNeil, B. J. The meaning and use of the area under a receiver operating characteristic (ROC) curve. *Radiology*, 143:29–36, 1982.

Hastie, T., Tibshirani, R., and Friedman, J. *The Elements of Statistical Learning: Data Mining, Inference, and Prediction*. Springer, 2001.

Koehn, Philipp, Och, Franz Josef, and Marcu, Daniel. Statistical phrase-based translation. In *Proc. of the Conf. of the North American Chapter of the Association for Computational Linguistics on Human Language Technology*, pp. 48–54, 2003.

Langley, Pat. The changing science of machine learning. *Machine Learning*, 82:275–279, 2011.

Lobo, Jorge M., Jimnez-Valverde, Alberto, and Real, Raimundo. AUC: a misleading measure of the performance of predictive distribution models. *Global Ecology and Biogeography*, 17(2):145–151, 2008.

SRI International. CALO: Cognitive assistant that learns and organizes. http://caloproject.sri.com, 2003–2009.

Student. The probable error of a mean. *Biometrika*, 6(1):1–25, 1908.

van Iterson, M., van Haagen, H.H.B.M., and Goeman, J.J. Resolving confusion of tongues in statistics and machine learning: A primer for biologists and bioinformaticians. *Proteomics*, 12:543–549, 2012.

van Rijsbergen, C. J. *Information Retrieval*. Butterworth, 2nd edition, 1979.

Warrick, P. A., Hamilton, E. F., Kearney, R. E., and Precup, D. A machine learning approach to the detection of fetal hypoxia during labor and delivery. In *Proc. of the Twenty-Second Innovative Applications of Artificial Intelligence Conf.*, pp. 1865–1870, 2010.

Zdziarski, Jonathan A. *Ending Spam: Bayesian Content Filtering and the Art of Statistical Language Classification*. No Starch Press, San Francisco, 2005.